\documentclass{article}

\usepackage[preprint]{neurips_2026}
\setcitestyle{numbers,square,comma}

\usepackage[utf8]{inputenc} 
\usepackage[T1]{fontenc}    
\usepackage{hyperref}       
\usepackage{url}            
\usepackage{booktabs}       
\usepackage{amsfonts}       
\usepackage{nicefrac}       
\usepackage{microtype}      
\usepackage{xcolor}         
\usepackage{multirow}
\usepackage{makecell}
\usepackage{graphicx}
\title{
Decomposing Queries into Tool Calls for Long-Video Keyframe Retrieval}

\author{
  Michal Shlapentokh-Rothman \quad Prachi Garg \quad Yu-Xiong Wang \quad Derek Hoiem \\
  University of Illinois at Urbana-Champaign \\
  \texttt{\{michal5, prachig3, yxw, dhoiem\}@illinois.edu}
}
\usepackage{listings}
\begin{document}

\maketitle

\begin{abstract}
Keyframe selection is a direct way to provide verifiable visual evidence for long-video question answering (QA). Queries differ in what they require, and finding the right frames depends on knowing what to look for. Existing keyframe selectors either score every frame against a single query, or decompose the query into a fixed schema evaluated by a single visual tool. We propose ToolMerge, a keyframe retrieval method based on decomposition and merging: an Large Language Model (LLM) based planner decomposes the query into tool calls and specifies how their per-tool rankings are merged using boolean operators. To evaluate retrieval directly, we construct Molmo-2 Moments (M2M), a benchmark in which every question is anchored to a specific time interval by construction. Across QA, question retrieval, and caption retrieval, ToolMerge is competitive with prior keyframe selectors, most notably on caption retrieval, outperforming other methods by 5\%. Code and data can be found at \url{https://github.com/michalsr/ToolMerge}.
\end{abstract}

\section{Introduction}

Verifying outputs of long-video systems is important for trustworthy human-AI interaction. Keyframes provide a natural source of visual evidence and can be passed directly to a vision-language model for answering.
Our goal is a long-video question answering (QA) system that is verifiable, correct, and efficient. We approach this with ToolMerge, a system that uses lightweight visual tools to find relevant keyframes across long videos and passes them to a vision-language model (VLM) for answering. Input queries vary widely in what they require: some specify a single scene, others multiple objects or entities. Existing keyframe selection approaches either score every frame against a single query, or decompose the query into a fixed schema of objects and relations. The key idea behind ToolMerge is decomposition and merging: we separate the query to leverage different tools that focus on different aspects of the video evidence, and merge their per-tool rankings using boolean operators. This is controlled automatically by an LLM-based planner, which can be executed zero-shot or further improved with RL post-training. The tools are lightweight vision models rather than per-frame captioners, enabling search across the full video at low cost.


%


Evaluating such a system requires a benchmark that isolates retrieval from answering. Noting a gap in existing benchmarks (see Section~\ref{sec:datasets}), we propose \textbf{Molmo-2 Moments (M2M)}, built from the Molmo-2 Captioning Dataset~\citep{molmo2}, which provides captions describing specific time intervals  within longer videos. Every M2M question is generated from one Molmo-2 caption, and is therefore anchored to a specific time interval by construction. M2M questions are multiple choice, and a multi-stage filtering pipeline addresses several aspects of question design, including ensuring each question is answerable from its time interval. The test and validation splits additionally pass through human verification, where questions humans answer incorrectly are removed. M2M supports two evaluations: question retrieval (whether selected frames for a given question fall in the ground truth interval) and downstream QA accuracy. A third evaluation, caption retrieval, runs directly on the Molmo-2 Captioning Dataset: the system must select frames corresponding to the caption. We evaluate on M2M and caption retrieval alongside two established long-video QA benchmarks, LongVideoBench~\citep{longvideobench} and Video-MME~\citep{videomme}, comparing against prior keyframe selectors and a strong single-tool retrieval baseline using multiple downstream VLMs and frame budgets.





We summarize our contributions as follows:

\begin{itemize}
\item ToolMerge, a planner-based method that decomposes queries into tool calls and merges their per-tool rankings using boolean operators, executable zero-shot or further improved with RL post-training.
\item Molmo-2 Moments (M2M), a benchmark in which every question is anchored to a specific time interval by construction.
\item A three-part evaluation across long-video QA, direct retrieval on M2M, and caption-based retrieval, with improvements over prior keyframe selectors most pronounced on caption-based retrieval.
\end{itemize}

\section{Related Work}


Below we discuss related work on methods for long-video question answering (with a focus on keyframe selection) and benchmarks for evaluating them.
\paragraph{Caption Based Long-Video QA} 
A common approach to long-video QA is to generate textual descriptions of the video and let an LLM reason over the output to answer the question. Descriptions are produced in a variety of ways: through dense short-clip captioning~\citep{llovi}, query-adaptive hierarchies~\citep{video_tree}, iterative retrieval and re-captioning~\citep{videoagent}, or planning-based perception with parallel reduction or evidence-seeking loops~\citep{mrvideo,dvd,avp}. While the strongest of these reach high accuracy, they require many sequential VLM and LLM calls per question and are expensive at inference time. We take a different approach, instead selecting a small number of keyframes (eg keyframe selection) first and generating the answer from those frames only, resulting in both reduced inference costs and visual evidence a user can inspect. 

\paragraph{Trained Keyframe Selection} Within keyframe-selection methods, one line of work learns how to select the best keyframes: Frame-Voyager~\citep{framevoyager} uses a ranking objective, ~\citet{mllm_frames} uses a vision-language model (VLM) as a scoring policy, and TimeSearch-R~\citep{timesearchr} uses reinforcement learning. Our core method is training-free and we additionally investigate whether the planning LLM can be improved with reinforcement learning given the same tool set.
\paragraph{Single-Query Keyframe Selection} One line of training-free keyframe selection scores every frame against the entire question using image-text similarity from a vision-language model, then applies a selection strategy over the per-frame scores. The methods share this scoring formulation and differ in their selection strategies: AKS~\citep{aks} balances relevance with explicit temporal coverage under a fixed frame budget, BOLT~\citep{bolt} uses inverse transform sampling, MDP3~\citep{mdp3} uses a variant of Determinantal Point Process (DPP) with a Markov decision process, Q-Frame~\citep{qframe} uses Gumbel-Max with multiple resolutions and one of the more recent methods, WFS-SB~\citep{wfssb}, uses wavelet transforms to segment the video and Maximal Marginal Relevance (MMR) to select frames within each segment. All of these methods score frames against a single query, typically the question concatenated with the answer choices. Our approach, like the methods discussed next, breaks the query into multiple parts that are searched for separately.

\paragraph{Decomposed-Query Keyframe Selection} In contrast to single-query methods, decomposed-query approaches first parse the question into structured components, usually different types of objects and search for each separately. T*~\citep{t_star} uses a VLM to extract target objects from the query and scores frames with YOLO-World~\citep{yolo_world}, applying adaptive spatial-temporal zoom-in to refine the candidate set across iterations. Logic-in-Frames (LIF)~\citep{lif} extends this with four predefined types of binary relations between objects, decomposing each query into a fixed schema of (object, relation) tuples scored by the object detector. Both methods iterate until either the target evidence is identified or a per-question frame budget is exhausted. We extend this direction along two axes. First, decomposition is not constrained to a fixed schema: the planner emits free-form text sub-queries rather than (object, relation) tuples, which lets a single query be decomposed into separate searches using different tools. Second, per-frame evidence is combined based on the rankings of independent tool calls in a single pass rather than within a budget-bounded iterative loop, and every frame is scored on every call.

\label{sec:datasets}
\paragraph{Video QA and Keyframe Datasets} Long-video QA benchmarks are not well-suited to measuring keyframe selection in isolation. Questions are often underspecified, with evidence that could come from multiple moments in the video. Distractor design is also hard: on Video-MME~\citep{videomme}, a blind model answers a substantial fraction of questions correctly without seeing the video. LV-Haystack~\citep{t_star} takes on the difficult task of annotating keyframes for questions drawn from existing benchmarks. Because those questions were not originally written with specific moments in mind, many have multiple valid evidence frames, and any single-keyframe annotation is inherently incomplete. We take the opposite starting point: we derive questions from detailed descriptions of specific video clips from Molmo2-Captioning~\citep{molmo2}, so each question is anchored to a specific moment by construction.

\section{Method}
\label{sec:method}

\begin{figure}
    \centering
    \includegraphics[width=1\linewidth]{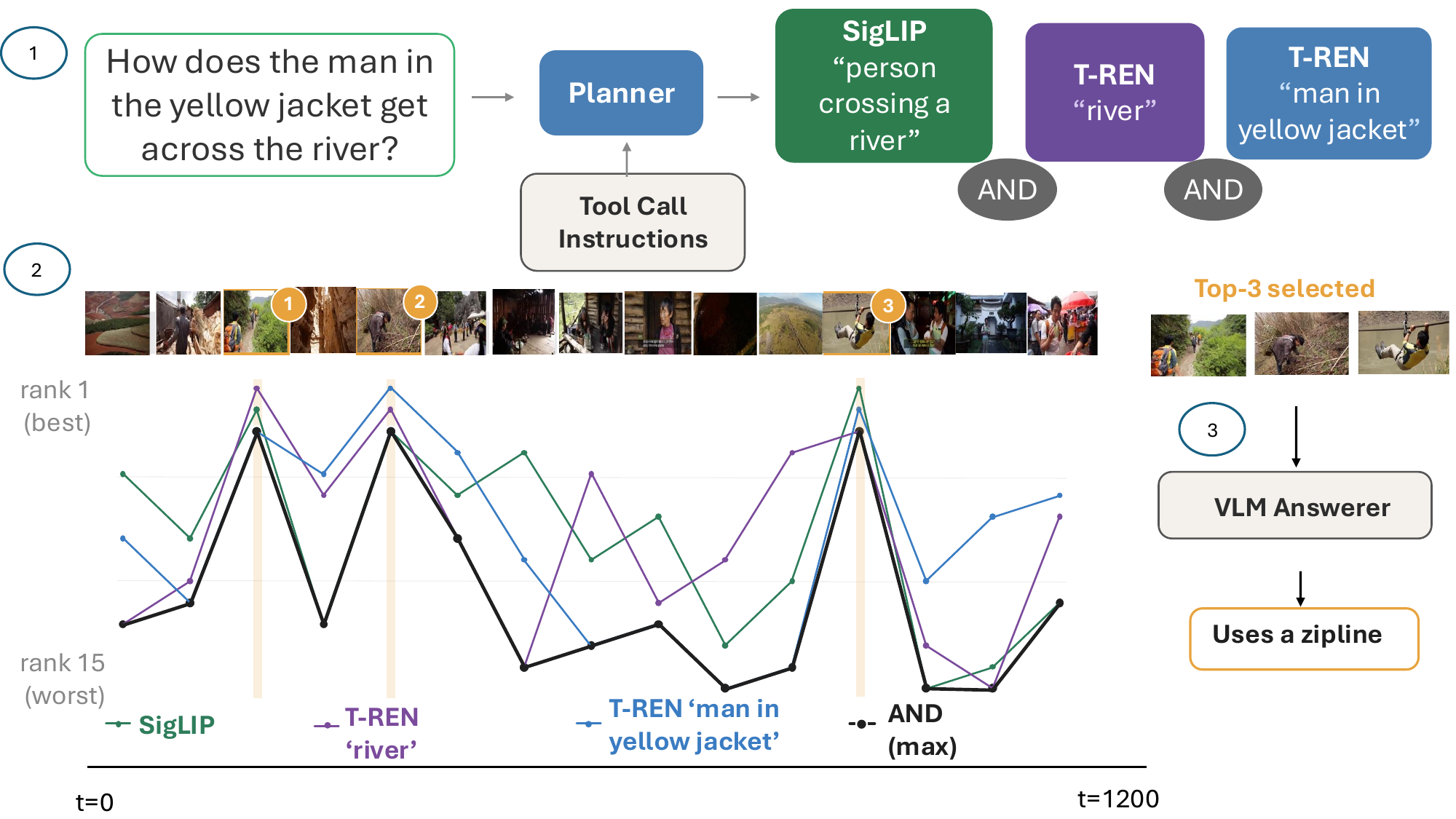}
    \caption{Overview of our main method, ToolMerge. (1) A text-only planner reads the question and answer choices and emits independent tool calls: SigLIP-2 for the action (`person crossing river') and T-REN for two entities (`man in yellow jacket', `river'). The planner combines all three calls under AND because every element must appear together in the answering frame (2) Each tool scores every frame, producing per-frame ranks (1 = best). The AND operator merges ranks by taking the worst rank across tools at each frame, so the combined ordering peaks where all required evidence appears together. (3) The top-k frames  are passed to a VLM (the `answerer') alongside the question.}
    \label{fig:main}
\end{figure}

\paragraph{Problem Definition} Given a video $V$ with $|V|$ frames and a query $q$ with answer choices $C$ (if relevant), query-conditioned keyframe selection aims to select a subset $V_k \subset V$ with $k \ll |V|$ such that a VLM conditioned on $V_k$ can correctly respond to $q$.

\paragraph{Method Overview} A central design question for long video keyframe extraction is how to extract evidence efficiently without using expensive tools like captioning models on every frame. Our method, ToolMerge, has three stages (see Figure~\ref{fig:main}) and addresses this by combining efficient visual tools under a lightweight planner. In the first stage, an LLM, referred to as the `planner', outputs 1) $N$ tool calls and 2) a set of boolean operators specifying how the tool call outputs should be combined. During the second stage, the tools are executed, and the outputs merged via the operators specified in the first stage. Lastly, the top-k frames are selected and passed to a vision-language model (VLM) for answering. The planner has access to two complementary tools: SigLIP-2~\citep{siglip2} scores frames via whole-frame image/text matching, capturing scene-level evidence. T-REN~\citep{tren} scores frames via text-aligned region features, contributing additional signal for queries where the relevant evidence is a localized entity rather than the overall scene. We additionally run OCR on every sampled frame for reading on-screen text.

\paragraph{Planner} Different queries require different tool combinations. Queries about small objects benefit from precise localization, while queries specifying multiple co-occurring visual elements require multiple tool calls combined under different operators. Given the input query with answer choices and tool descriptions, the planner decides how many times to call SigLIP-2 and T-REN  (if any), the corresponding text inputs, and how to combine the outputs. The input prompt can be found in Appendix~\ref{app:prompt-planner}.

\begin{figure}
    \centering
    \includegraphics[width=.8\linewidth]{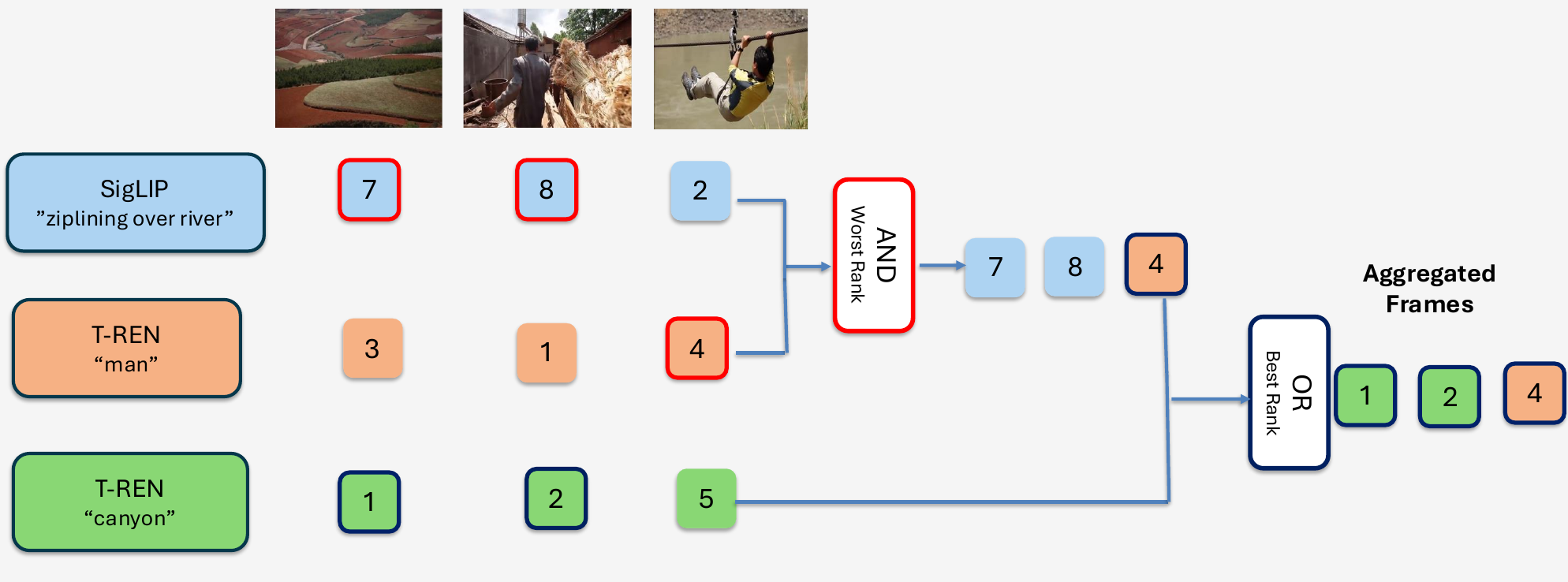}
    \caption{Merging Example. Each frame has a rank per tool call. AND operators choose the worst rank and OR operators select the best rank. At the end, each frame has a single rank.}
    \label{fig:rank}
\end{figure}

\paragraph{Merging} 
Each tool call is executed on the same set of frames, resulting in 
$N$ scores per frame. Our goal is to combine these into a single ranking over the frames.  However, since individual tools have different scoring mechanisms, we cannot directly compare scores across tools. Instead, each frame is assigned a rank within each tool call, eg the frame with the highest score for tool call 1 is given rank 1. The AND and OR operators then combine ranks from pairs of tool calls into a single score: AND selects the maximum rank (worse) while OR selects the minimum rank (better). When more than two tool calls are present operators are applied left to right as specified by the planner. See Figure~\ref{fig:rank} for an example. 

Separately, OCR contributes frames directly to the merged ranking. After OCR is run on each sampled frame, near-duplicate extractions from adjacent frames are removed via fuzzy matching, and a cheap LLM judge (GPT-4o-Mini) filters the remaining extractions for relevance to the query. Frames whose text survives this filter are temporally grouped using a window of $\tau$ seconds, and the median frame of each group is kept. Each kept frame is inserted at rank 1 in the merged ranking, giving OCR frames priority over SigLIP-2 and T-REN frames since the LLM judge has already confirmed query relevance, so these frames carry less ambiguity than frames selected by visual similarity alone.

\paragraph{Final Frame Selection}
From the resulting ranking, we perform greedy NMS~\citep{mllm_frames} where we select 
up to $k$ frames greedily, in order of merged rank (best first), keeping a frame only 
if it is at least $\tau$ seconds from every already-selected frame. The 
selected frames are sorted in temporal order before being passed to the 
answerer.

\section{Benchmark Construction}
\label{sec:dataset_ant}

We want video-question pairs where (1) the answer is unambiguous, (2) the question is correctly answerable from the relevant clip, and (3) the question is difficult to answer without seeing the video or by sampling a few uniform frames. The start and end times of the source clip serve as ground-truth for both retrieval and QA evaluation.
We construct a benchmark for evaluating keyframe selection by using the Molmo2-Captioning dataset~\citep{molmo2}, in which annotators verbally describe short segments referred to as clips (typically 10–20 seconds) of longer videos. Everything described in a caption is assumed to be visible within the clip's start and end times.

Starting with the video clip caption, an LLM generates candidate questions, which then pass through a sequence of filters and rewrites for a total of 8 steps. The goal is for each surviving question to be answerable from its source clip and not from anywhere else. The number of questions remaining after each step can be seen in Table~\ref{tab:filter-counts}. Prompts used for dataset generation can be found in Appendix~\ref{app:prompts}.
\begin{enumerate}
    \item \textbf{Question generation}: For each clip, an LLM (GPT-5.4) generates candidate 5-option multiple-choice questions from the clip's caption, optionally augmented with a summary of other clips from the same video. The number of candidates per clip is uncapped.
    \item \textbf{Blind-LLM filter}: To verify that questions require visual evidence, we discard questions that an LLM (GPT-5.2) can answer without looking at any frames from the video. 
    \item \textbf{Answer choice rewrite}: Some questions fail step 2 not because they are genuinely answerable from text but because the correct answer is the only plausible option among the other answer choices. To recover these, we ask an LLM (GPT-5.2) to propose alternative answer choices for each discarded question and re-run the blind filter. Questions where the blind model now fails are kept.
    \item \textbf{Scope filter}: Given the question, its caption, and the optional video summary, an LLM (GPT-5.2) removes questions that cannot be verified visually within the clip, including questions about audio or narration, or questions whose evidence lies outside what the caption describes. The LLM also removes questions phrased in a way that reveals the benchmark's construction (eg, references to the clip or caption), since questions should assume the viewer sees the entire video.
    \item \textbf{Cleanup rewrite}: Another filtering pass resolves recurring issues in surviving questions. Given the question, source caption, and optional summary, an LLM (GPT-5.2) applies a several rewrites: underspecified questions gain visual detail from the caption, overly similar answer choices are made more distinct, and proper names (e.g., `Steve') are replaced with visual descriptions. 
    \item \textbf{Necessity Filter}: All previous steps operated on text only. We then remove any question answered correctly by a VLM (GPT-4.1-Mini) given 8 frames uniformly sampled from the full video, since such questions do not require targeted retrieval.
    \item \textbf{Difficulty Filter}: Next, questions answered incorrectly given 16 uniformly sampled frames from inside the ground-truth clip are removed. Since the goal is to evaluate retrieval, questions that cannot be answered even from the right frames are bottlenecked by reasoning and excluded. A question is discarded only if both GPT-5.2 and GPT-4.1-Mini answer it incorrectly to avoid biasing the questions towards one particular LLM ($\approx$ 20\% of questions GPT-5.2 gets wrong are answered correctly by GPT-4.1-Mini).
    \item \textbf{Diversity Sampling}: To balance per-clip coverage, an LLM selects at most 4 questions per clip. 
    
\end{enumerate}
This yields 11,673 questions split into 9,677 train, 997 validation, and 999 test, with no overlapping videos across splits. Videos range from 3 minutes to 4 hours, with an average time of 19.3 minutes. For the validation and test splits, we add a human-verification pass on top of the automated filters: each question is answered by one of three annotators after viewing the ground-truth clip, and questions answered incorrectly are dropped. This pass catches failure modes outside the reach of the automated filters, including questions grounded in inaccurate captions and questions about niche content that a VLM may recognize but a typical viewer would not. This leaves 756 test questions and 748 validation questions. 


\begin{table}[h]
\centering
\begin{tabular}{clrr}
\toprule
Step & Description & Questions Remaining & \% of Previous \\
\midrule
1    & After question generation          & 156,595 & --- \\
2--3 & After blind filter (with recovery) &  98,286 & 62.8\% \\
4    & After scope filter                 &  58,455 & 59.5\% \\
5    & After cleanup rewrite              &  58,455 & 100\% \\
6    & After necessity filter             &  22,969 & 39.3\% \\
7    & After difficulty filter            &  14,808 & 64.5\% \\
8    & After diversity sampling             &  11,673 & 78.8\% \\
\bottomrule
\end{tabular}
\caption{Question counts after each step in the benchmark construction pipeline. Steps 2 and 3 are reported jointly because step 3 (answer-choice rewrite) recovers questions discarded by step 2 (blind filter).}
\label{tab:filter-counts}
\end{table}

\section{Experiments}

\label{sec:exp}


\paragraph{Implementation Details}
ToolMerge uses SigLIP-2-Giant~\citep{siglip2} for image-text matching and T-REN for region-text matching, with OCR handled by EasyOCR~\citep{easyocr} paired with GPT-4o-Mini (accessed through Azure) as an LLM judge. Qwen3-VL-8B~\citep{qwen3vl} serves as the planner across all experiments, and runs with zero input frames. The grouping parameter $\tau$, used both to cluster frames during OCR and for greedy NMS~\citep{mllm_frames} (e.g. minimal temporal gap), is set to $\min(\frac{D }{2K}, 10)$
seconds, where $D$ is the full video duration (not the clip length) and $K$ is the number of selected frames.

We compare against three baselines: blind inference (no frames), uniform sampling, and a top-K baseline that retrieves frames most similar to the concatenated question and answer choices using the same SigLIP-2 model, selected with greedy NMS, referred to as SigLIP-Q. Beyond these, we evaluate recent keyframe methods spanning both single-query and decomposition-based designs. For a fair comparison, we evaluate every additional method using released code under matched settings: temperature 0, SigLIP-2-Giant as the image-text matcher (for BOLT~\citep{bolt}, WFS~\citep{wfssb} and AKS~\citep{aks}), Qwen3-VL~\citep{qwen3vl} for planning (for LIF~\citep{lif}) and video sampling (for image/text matching) at 2 FPS. All other hyperparameters remain at their defaults. Additional comparisons appear in Appendix~\ref{sec:q25}. Code and dataset annotations will be publicly released. 

We use two different answering models: Qwen3-VL-8B and GPT-4o~\citep{gpt4o} with two  different values of K: 8 and 32 following LIF.

\subsection{Long Video QA on Existing Benchmarks}
\label{tab:vqa}

\begin{table*}[!htb]
\centering
\begingroup
\renewcommand{\arraystretch}{1.12}
\setlength{\tabcolsep}{3pt}

\def\LVBDUp#1{{\scriptsize\textcolor{green!45!black}{(+#1)}}}
\def\LVBDDown#1{{\scriptsize\textcolor{red!70!black}{(-#1)}}}

\def\LVBDRes#1#2{%
  \makebox[2.25em][r]{#1}\,\makebox[2.75em][l]{#2}%
}

\resizebox{\linewidth}{!}{%
\begin{tabular}{lcccccccc}
\toprule
Method
& \multicolumn{4}{c}{Qwen3-VL}
& \multicolumn{4}{c}{GPT-4o} \\
\cmidrule(lr){2-5}
\cmidrule(lr){6-9}

& \multicolumn{2}{c}{LongVideoBench}
& \multicolumn{2}{c}{Video-MME}
& \multicolumn{2}{c}{LongVideoBench}
& \multicolumn{2}{c}{Video-MME} \\
\cmidrule(lr){2-3}
\cmidrule(lr){4-5}
\cmidrule(lr){6-7}
\cmidrule(lr){8-9}

& 8 & 32
& 8 & 32
& 8 & 32
& 8 & 32 \\

\midrule

Blind Text (0 frames)
& \multicolumn{2}{c}{42.6}
& \multicolumn{2}{c}{44.5}
& \multicolumn{2}{c}{38.0}
& \multicolumn{2}{c}{38.7} \\

\midrule

Uniform
& \LVBDRes{58.2\hphantom{*}}{} & \LVBDRes{63.7\hphantom{*}}{}
& \LVBDRes{57.9\hphantom{*}}{} & \LVBDRes{67.3\hphantom{*}}{}
& \LVBDRes{56.2\hphantom{*}}{} & \LVBDRes{64.4\hphantom{*}}{}
& \LVBDRes{64.0\hphantom{*}}{} & \LVBDRes{70.6\hphantom{*}}{} \\

SigLIP-Q
& \LVBDRes{60.8\hphantom{*}}{\LVBDUp{2.6}} & \LVBDRes{65.1\hphantom{*}}{\LVBDUp{1.4}}
& \LVBDRes{58.5\hphantom{*}}{\LVBDUp{0.6}} & \LVBDRes{68.0\hphantom{*}}{\LVBDUp{0.7}}
& \LVBDRes{59.8\hphantom{*}}{\LVBDUp{3.6}} & \LVBDRes{63.9\hphantom{*}}{\LVBDDown{0.5}}
& \LVBDRes{64.2\hphantom{*}}{\LVBDUp{0.2}} & \LVBDRes{69.2\hphantom{*}}{\LVBDDown{1.4}} \\

AKS (CVPR 2025)
& \LVBDRes{56.7\hphantom{*}}{\LVBDDown{1.5}} & \LVBDRes{63.3\hphantom{*}}{\LVBDDown{0.4}}
& \LVBDRes{58.5\hphantom{*}}{\LVBDUp{0.6}} & \LVBDRes{67.3\hphantom{*}}{\LVBDUp{0.0}}
& \LVBDRes{56.0\hphantom{*}}{\LVBDDown{0.2}} & \LVBDRes{62.3\hphantom{*}}{\LVBDDown{2.1}}
& \LVBDRes{64.7\hphantom{*}}{\LVBDUp{0.7}} & \LVBDRes{72.5\hphantom{*}}{\LVBDUp{1.9}} \\

BOLT (CVPR 2025)
& \LVBDRes{58.0\hphantom{*}}{\LVBDDown{0.2}} & \LVBDRes{65.0\hphantom{*}}{\LVBDUp{1.3}}
& \LVBDRes{62.4\hphantom{*}}{\LVBDUp{4.5}} & \LVBDRes{67.4\hphantom{*}}{\LVBDUp{0.1}}
& \LVBDRes{56.6\hphantom{*}}{\LVBDUp{0.4}} & \LVBDRes{63.2\hphantom{*}}{\LVBDDown{1.2}}
& \LVBDRes{67.0\hphantom{*}}{\LVBDUp{3.0}} & \LVBDRes{71.7\hphantom{*}}{\LVBDUp{1.1}} \\

LIF (NeurIPS 2025)
& \LVBDRes{55.3\hphantom{*}}{\LVBDDown{2.9}} & \LVBDRes{62.8\hphantom{*}}{\LVBDDown{0.9}}
& \LVBDRes{57.0\hphantom{*}}{\LVBDDown{0.9}} & \LVBDRes{62.8\hphantom{*}}{\LVBDDown{4.5}}
& \LVBDRes{56.1\hphantom{*}}{\LVBDDown{0.1}} & \LVBDRes{62.8\hphantom{*}}{\LVBDDown{1.6}}
& \LVBDRes{63.6\hphantom{*}}{\LVBDDown{0.4}} & \LVBDRes{67.2\hphantom{*}}{\LVBDDown{3.4}} \\

WFS (CVPR 2026)
& \LVBDRes{60.4\hphantom{*}}{\LVBDUp{2.2}} & \LVBDRes{64.9\hphantom{*}}{\LVBDUp{1.2}}
& \LVBDRes{63.3\hphantom{*}}{\LVBDUp{5.4}} & \LVBDRes{67.5\hphantom{*}}{\LVBDUp{0.2}}
& \LVBDRes{59.6\hphantom{*}}{\LVBDUp{3.4}} & \LVBDRes{64.5\hphantom{*}}{\LVBDUp{0.1}}
& \LVBDRes{66.6\hphantom{*}}{\LVBDUp{2.6}} & \LVBDRes{72.6\hphantom{*}}{\LVBDUp{2.0}} \\

ToolMerge
& \LVBDRes{61.8\hphantom{*}}{\LVBDUp{3.6}} & \LVBDRes{\textbf{67.4}*}{\LVBDUp{3.7}}
& \LVBDRes{64.6\hphantom{*}}{\LVBDUp{6.7}} & \LVBDRes{\textbf{70.6}*}{\LVBDUp{3.3}}
& \LVBDRes{61.3\hphantom{*}}{\LVBDUp{5.1}} & \LVBDRes{65.3\hphantom{*}}{\LVBDUp{0.9}}
& \LVBDRes{\textbf{71.0}*}{\LVBDUp{7.0}} & \LVBDRes{73.2\hphantom{*}}{\LVBDUp{2.6}} \\

\midrule

ToolMerge + GRPO
& \LVBDRes{\textbf{63.6}\hphantom{*}}{\LVBDUp{5.4}} & \LVBDRes{66.8\hphantom{*}}{\LVBDUp{3.1}}
& \LVBDRes{\textbf{64.7}\hphantom{*}}{\LVBDUp{6.8}} & \LVBDRes{70.2\hphantom{*}}{\LVBDUp{2.9}}
& \LVBDRes{\textbf{63.5}\hphantom{*}}{\LVBDUp{7.3}} & \LVBDRes{\textbf{65.4}\hphantom{*}}{\LVBDUp{1.0}}
& \LVBDRes{70.7\hphantom{*}}{\LVBDUp{6.7}} & \LVBDRes{\textbf{73.7}\hphantom{*}}{\LVBDUp{3.1}} \\

\bottomrule
\end{tabular}%
}

\endgroup
\caption{Long-video QA accuracy on LongVideoBench and Video-MME, with Qwen3-VL or GPT-4o as the answerer at different frame budgets. ToolMerge + GRPO improves over the Uniform baseline across all settings and gives the highest accuracy in most groups. The single-query SigLIP-Q baseline is surprisingly competitive and LIF, the prior decomposition-based method, underperforms. \textbf{Highest in group}. * denotes statistical significance.}
\label{tab:lvb_videomme}
\end{table*}
In our first set of experiments, we look at the effectiveness of our planning method compared to other keyframe selection methods and baselines. We evaluate on two standard long video benchmarks, LongVideoBench~\citep{longvideobench} and VideoMME~\citep{videomme} without subtitles on either and using accuracy as the main metric. The results are shown in Table~\ref{tab:lvb_videomme}. As seen in the table, our method outperforms all other keyframe selectors. Notably, the SigLIP-Q baseline (top-k selection with a temporal gap on per-frame SigLIP-2 similarity scores) is a strong baseline despite its simplicity, achieving competitive performance with the more elaborate keyframe selection methods. LIF, the prior decomposition-based method, underperforms uniform sampling frequently, likely due to its limited search.
\subsection{Molmo2-Moment Question and Retrieval Evaluation}
\label{sec:m2m}
We evaluate both retrieval and QA accuracy on our Molmo2-Moment (M2M) dataset constructed in Section~\ref{sec:dataset_ant}. 

\paragraph{Benchmark Validation} Before comparing methods on M2M, we verify that our benchmark satisfies its design goals. In Table~\ref{tab:main_results_ours}, both blind-text and uniform-sampling accuracy on M2M are substantially lower than on existing long-video benchmarks. Blind text reaches 29.8\% and 29.9\% with both Qwen3-VL and GPT-4o respectively (chance is 20\% for 5-way MCQ), compared to 44.5\% (Qwen3-VL) and 38.7\% (GPT-4o) on Video-MME. Uniform sampling shows the same pattern at 8 frames: 35.8\% (Qwen3-VL) and 39.3\% (GPT-4o). Oracle performance, where frames are sampled from inside the ground-truth clip, reaches 77.9\% with GPT-4o at 8 frames and 79.4\% at 32 frames, indicating that retrieval and QA accuracy are closely connected.



\paragraph{Molmo2-Moment Question Retrieval and QA} We first evaluate the question retrieval ability of different methods. We measure HIT@$K$, the fraction of questions (in M2M) where at least one of the top-$K$ retrieved frames falls inside the ground-truth clip.Results are in Table~\ref{tab:question_retrieval}. Surprisingly, SigLIP-Q performs strongly, on par with ToolMerge at $K=8$ with Qwen3-VL and slightly higher at $K=32$. Such results indicate that on visually direct queries, basic methods like SigLIP-Q can be sufficient. Other keyframe selection methods underperform both our method and SigLIP-Q.  Because M2M questions
are answerable from the right frames without additional reasoning,  retrieval quality should map cleanly onto downstream QA accuracy.  We can confirm this by looking at downstream QA accuracy in Table~\ref{tab:main_results_ours}, where a similar pattern appears. 
\begin{table}[htbp]
\centering
\small
\begingroup
\setlength{\tabcolsep}{6pt}
\def\OURUp#1{{\scriptsize\textcolor{green!45!black}{(+#1)}}}
\def\OURDown#1{{\scriptsize\textcolor{red!70!black}{(-#1)}}}
\def\OURRes#1#2{%
  \makebox[2.25em][r]{#1}\,\makebox[2.75em][l]{#2}%
}
\begin{tabular}{lcccc}
\toprule
Method
& \multicolumn{2}{c}{Qwen3-VL}
& \multicolumn{2}{c}{GPT-4o} \\
\cmidrule(lr){2-3}
\cmidrule(lr){4-5}
& 8 & 32
& 8 & 32 \\
\midrule
Blind Text (0 frames)
& \multicolumn{2}{c}{29.8}
& \multicolumn{2}{c}{29.9} \\
\midrule
Uniform
& \OURRes{35.9}{} & \OURRes{51.9}{}
& \OURRes{39.3}{} & \OURRes{53.3}{} \\
Oracle
& \OURRes{68.0}{\OURUp{32.1}} & \OURRes{76.9}{\OURUp{25.0}}
& \OURRes{77.9}{\OURUp{38.6}} & \OURRes{79.4}{\OURUp{26.1}} \\
\midrule
SigLIP-Q
& \OURRes{61.6}{\OURUp{25.7}} & \OURRes{63.1}{\OURUp{11.2}}
& \OURRes{57.6}{\OURUp{18.3}} & \OURRes{61.1}{\OURUp{7.8}} \\
AKS (CVPR 25)
& \OURRes{57.5}{\OURUp{21.6}} & \OURRes{60.3}{\OURUp{8.4}}
& \OURRes{55.0}{\OURUp{15.7}} & \OURRes{58.2}{\OURUp{4.9}} \\
BOLT (CVPR 25)
& \OURRes{52.5}{\OURUp{16.6}} & \OURRes{58.2}{\OURUp{6.3}}
& \OURRes{52.2}{\OURUp{12.9}} & \OURRes{59.5}{\OURUp{6.2}} \\
LIF (NeurIPS 25)
& \OURRes{45.0}{\OURUp{9.1}} & \OURRes{51.9}{}
& \OURRes{51.5}{\OURUp{12.2}} & \OURRes{52.8}{\OURDown{0.5}} \\
WFS (CVPR 26)
& \OURRes{57.5}{\OURUp{21.6}} & \OURRes{59.0}{\OURUp{7.1}}
& \OURRes{56.8}{\OURUp{17.5}} & \OURRes{58.2}{\OURUp{4.9}} \\
ToolMerge
& \OURRes{61.6}{\OURUp{25.7}} & \OURRes{62.7}{\OURUp{10.8}}
& \OURRes{60.8}{\OURUp{21.5}} & \OURRes{60.7}{\OURUp{7.4}} \\
\midrule 
ToolMerge + GRPO 
& \OURRes{\textbf{63.0}}{\OURUp{{27.1}}} & \OURRes{\textbf{63.4}}{\OURUp{11.5}}
& \OURRes{\textbf{62.3}}{\OURUp{23.0}} & \OURRes{\textbf{62.3}}{\OURUp{10.0}} \\
\bottomrule
\end{tabular}
\endgroup
\caption{Downstream QA accuracy on Molmo2-Moment with Qwen3-VL or GPT-4o as the answerer for different frame budgets.  Oracle exceeds Uniform by roughly 30 points at $K=8$, confirming that retrieval is strongly connected to QA on M2M. ToolMerge with GRPO outperforms all other methods  and SigLIP-Q has surprisingly strong performance.  \textbf{Highest per group}. }
\label{tab:main_results_ours}
\end{table}

\begin{table}[htbp]
\centering
\small
\setlength{\tabcolsep}{6pt}
\begin{tabular}{lcccccc}
\toprule
Method & HIT@1 & HIT@2 & HIT@4 & HIT@8 & HIT@16 & HIT@32 \\
\midrule
Uniform          & 2.8  & 3.4  & 8.7  & 17.1 & 29.2 & 42.9 \\
SigLIP-Q         & \textbf{22.0} & \textbf{31.6} & 46.6 & 59.8 & \textbf{75.7} & \textbf{88.0} \\
AKS (CVPR 25)    & 0.0  & 28.2 & 35.6 & 42.9 & 55.8 & 74.2 \\
BOLT (CVPR 25)   & 0.8  & 7.3  & 18.5 & 34.7 & 56.5 & 74.5 \\
LIF (NeurIPS 25) & 6.9  & 12.0 & 18.1 & 24.5 & 34.1 & 45.0 \\
WFS (CVPR 26)    & 16.9 & 20.8 & 29.9 & 44.3 & 43.5 & 60.7 \\
ToolMerge             & 19.2 & 30.7 & \textbf{46.9} & \textbf{63.6} & 74.7 & 85.5 \\
\midrule 
ToolMerge + GRPO      &  21.8 & 35.1 & 53.7         & 66.3         & 78.4 & 88.2  \\
\bottomrule
\end{tabular}
\caption{\textbf{Molmo2-Moment Question Retrieval Results.} HIT@$k$ reported as percentages. SigLIP-Q and ToolMerge are the top performing (with SigLIP-Q taking a slight edge) while the rest of the methods underperform. \textbf{Highest per group}.}
\label{tab:question_retrieval}
\end{table}
\paragraph{Caption Retrieval} We additionally evaluate retrieval on the same clip captions from which the Molmo2-Moment questions were generated. Compared to the questions, captions are substantially longer and specify multiple co-occurring visual elements. Table~\ref{tab:caption_example} shows caption excerpts alongside a question derived from it. Taking the captions corresponding to test-set questions yields 522 unique captions, since multiple questions can come from a single caption.  We sample 478 additional captions from clips not used in
training, validation, or test, for a total of 1000. 
 Table~\ref{tab:caption_retrieval} reports HIT@$K$ on this setting. ToolMerge's retrieval performance is comparable to its question-retrieval performance, while SigLIP-Q, LIF, AKS, and WFS all drop relative to their numbers on question retrieval. A possible explanation is that as the input query becomes more visually descriptive, the benefit of multi-tool becomes more apparent and using separate tools finds frames that single-query methods miss.
\begin{table}[htbp]
\centering
\small
\setlength{\tabcolsep}{6pt}
\begin{tabular}{lcccccc}
\toprule
Method & HIT@1 & HIT@2 & HIT@4 & HIT@8 & HIT@16 & HIT@32 \\
\midrule
Uniform          & 4.0  & 5.3  & 11.5 & 18.3 & 37.6 & 61.7 \\
SigLIP-Q         & \textbf{17.3} & 26.0 & 38.4 & 52.7 & 67.9 & 80.5 \\
AKS (CVPR 25)    & 0.0  & 21.7 & 27.9 & 36.4 & 50.7 & 65.9 \\
BOLT (CVPR 25)   & 1.1  & 4.4  & 15.5 & 29.1 & 47.8 & 71.9 \\
LIF (NeurIPS 25) & 4.0  & 7.0  & 13.0 & 20.1 & 27.4 & 36.5 \\
WFS (CVPR 26)    & 11.4 & 18.6 & 27.0 & 35.7 & 42.4 & 52.6 \\
ToolMerge             & 14.7 & \textbf{29.4} & \textbf{43.4}* & \textbf{58.1}* & \textbf{73.6}* & \textbf{85.3}* \\
\midrule 
ToolMerge + GRPO      & 15.9 & 33.6          & 48.8         &   61.5  & 75.1 & 85.6 \\ 
\bottomrule
\end{tabular}
\caption{Caption retrieval results, organized by method type. HIT@$k$ measures whether a relevant caption is retrieved within the top $k$ results. Unlike question retrieval and accuracy, our method outperforms SigLIP-Q at $k>1$. Similar to question retrieval, other keyframe methods underperform SigLIP-Q, most notably LIF. \textbf{Highest per group}. *statistically significant (if any)}
\label{tab:caption_retrieval}
\end{table}


\newcommand{\evidence}[1]{\textbf{\textcolor{blue!60!black}{#1}}}

\begin{table}[htbp]
\centering
\scriptsize
\setlength{\tabcolsep}{3pt}
\renewcommand{\arraystretch}{0.98}

\begin{tabular}{@{}>{\raggedright\arraybackslash}p{0.48\columnwidth}
                >{\raggedright\arraybackslash}p{0.48\columnwidth}@{}}
\toprule
Caption Excerpt & Generated Question \\
\midrule

\ldots Nearby instruments include \evidence{a fuel gauge located in the top right corner, which indicates the tank is three-quarters full}, and \evidence{an oil pressure gauge in the top left, showing a reading of approximately 60}. \ldots
&
When the top-left oil pressure gauge reads around 60, what does the top-right gauge show at the same time? \\

\addlinespace[2pt]

\ldots a pair of hands shaping copper-colored wire into a ring \ldots
\evidence{On one part of the wire bundle, there is a thin section
of copper wire mesh.} \ldots
&
When the hands shape the copper-colored wire into a ring, what additional detail is visible on part of the wire bundle? \\

\bottomrule
\end{tabular}

\caption{Examples of generated questions from caption excerpts. Colored text corresponds to information used to generate each question.}
\label{tab:caption_example}
\end{table}

\paragraph{Improving the Planner} All planner results so far are zero-shot with Qwen3-VL-8B. We ask how much can be improved by training the open-source planner on M2M, or by replacing it with a stronger commercial model without training. 
Without known-optimal tool-call sequences to supervise against, SFT on examples where the zero-shot planner produced correct answers does not improve over the zero-shot baseline. GRPO~\citep{grpo} is a more natural fit, since it sidesteps the supervision problem: each rollout's ranking can be scored directly against the ground-truth interval. For GRPO training (see Appendix~\ref{sec:training_details} for details), we filter the training split to questions the answerer gets wrong when given 8 frames uniformly sampled from the ground-truth interval, and we keep one question per clip, leaving 50\% of the training split. We train with a normalized recall reward: the number of selected frames whose timestamps fall inside the ground-truth interval, divided by the maximum number of frames that can fit inside the interval under greedy NMS.The reward avoids running the answerer during rollouts, saving memory and time while still transfering to downstream QA accuracy. This is consistent with M2M's design goal of separating retrieval from answering.
GRPO improves the planner across all three M2M evaluations as can be seen in  Tables~\ref{tab:main_results_ours},~\ref{tab:question_retrieval},~\ref{tab:caption_retrieval}. The largest improvements are on caption and question retrieval. Replacing Qwen3-VL-8B with GPT-5.4-Pro at zero shot also gives a comparable improvement: 8-frame M2M QA reaches 63.4 and HIT@8 on question retrieval reaches 67.0. This suggests the 8B planner can be further improved, especially for retrieval.

\paragraph{Time Per Frame/Question} Part of our motivation is to design a system that avoids the expensive process of captioning different frames. In Table~\ref{tab:timing_breakdown}, we show how long each component of our method takes for a video at 1 FPS  with one SigLIP-2 and one T-REN query on a single 40 GB A100. The results show the relative efficiency of our method compared to a captioning based approach, like DVD~\citep{dvd}. 

\label{sec:timing}

\begin{table*}[htbp]
\centering
\small
\setlength{\tabcolsep}{4pt}
\begin{tabular}{cccc cccc c}
\toprule
\multicolumn{4}{c}{Visual Pre-process (s)}
& \multicolumn{4}{c}{Query-Based (s)}
& \multicolumn{1}{c}{Question Answering(s)} \\
\cmidrule(lr){1-4}
\cmidrule(lr){5-8}
\cmidrule(lr){9-9}
SigLIP & T-REN & OCR & Captioning
& Plan Generation & T-REN & SigLIP & OCR 
& VLM \\
\midrule
52 & 44 & 30 & 428 
& 1.67 & .023 & .029 & 16 
& .45 \\
\bottomrule
\end{tabular}
\caption{Different time components for our method and time for captioning given a 10 minute video at 1 FPS with 1 SigLIP tool call and 1 T-REN tool call. Captioning the entire video takes 3.4x as much time as our pre-processing approach.}
\label{tab:timing_breakdown}
\end{table*}




\paragraph{Ablations} We test three design choices: which tools to use, how to combine their scores, and the temporal-gap parameter $\tau$. 
Table~\ref{tab:tool_ablation} reports performance with T-REN and OCR removed, regenerating the planner prompt for each tool combination. The effect depends on the dataset: removing OCR costs the most on the subtitle-heavy Video-MME, while removing T-REN costs the most on LongVideoBench, which is more visually grounded. On average across the three datasets, OCR contributes 1.7 points and T-REN is roughly neutral, indicating that T-REN's contribution is dataset-dependent rather than uniform.  Table~\ref{tab:method_ablation} compares rank-based merging against using raw tool scores directly, and against restricting the planner to a single tool call. Raw scores match rank merging on average, while restricting to a single tool call costs 1.0 points, indicating that using multiple tool calls matters more than score normalization. Lastly, Table~\ref{tab:tau} shows values of $\tau$ at 2,5, and 10 seconds. The optimal value differs across datasets: while M2M favors smaller $\tau$, LongVideoBench and Video-MME are best at $\tau = 10$, which is the default we use. 




\newcommand{\toolcheck}{\checkmark}

\begin{table*}[htbp]
\centering
\small
\begingroup
\setlength{\tabcolsep}{6pt}

\begin{tabular}{lccc cccccc@{\hspace{12pt}}cc}
\toprule
Method
& OCR & T-REN & SigLIP
& \multicolumn{2}{c}{M2M (Val)}
& \multicolumn{2}{c}{Video-MME}
& \multicolumn{2}{c}{LongVideoBench}
& Avg.
& $\Delta$ Avg. \\
\cmidrule(lr){2-4}
\cmidrule(lr){5-6}
\cmidrule(lr){7-8}
\cmidrule(lr){9-10}
\cmidrule(lr){11-11}
\cmidrule(lr){12-12}
&
&
&
& 8 & 32
& 8 & 32
& 8 & 32
& All
& vs. Ours \\
\midrule

Ours
& \toolcheck & \toolcheck & \toolcheck
& 62.5 & 60.8
& 64.6 & 70.6
& 61.8 & 67.4
& 64.6
& -- \\

w/o T-REN
& \toolcheck &  & \toolcheck
& 62.7 & 62.4
& 65.3 & 70.0
& 60.6 & 65.9
& 64.5
& $-0.1$ \\

w/o OCR
&  & \toolcheck & \toolcheck
& 60.4 & 60.2
& 62.3 & 66.7
& 61.6 & 66.5
& 63.0
& $-1.7$ \\

\bottomrule
\end{tabular}

\endgroup

\caption{\textbf{Tool ablation results} OCR has the largest average effect, while removing T-REN has a smaller overall effect.}
\label{tab:tool_ablation}
\end{table*}

\begin{table*}[htbp]
\centering
\small
\begingroup
\setlength{\tabcolsep}{6pt}

\begin{tabular}{lcccccc@{\hspace{12pt}}cc}
\toprule
Method
& \multicolumn{2}{c}{M2M (Val)}
& \multicolumn{2}{c}{Video-MME}
& \multicolumn{2}{c}{LongVideoBench}
& Avg.
& $\Delta$ Avg. \\
\cmidrule(lr){2-3}
\cmidrule(lr){4-5}
\cmidrule(lr){6-7}
\cmidrule(lr){8-8}
\cmidrule(lr){9-9}
& 8 & 32
& 8 & 32
& 8 & 32
& All
& vs. ToolMerge \\
\midrule

ToolMerge
& 62.5 & 60.8
& 64.6 & 70.6
& 61.8 & 67.4
& \textbf{64.6}
& -- \\

Rawscore
& 61.7 & 63.0
& 64.9 & 70.7
& 61.8 & 65.1
& 64.5
& $-0.1$ \\

Single Call
& 59.8 & 63.0
& 63.6 & 69.2
& 61.3 & 65.1
& 63.7
& $-1.0$ \\

\bottomrule
\end{tabular}

\endgroup

\caption{\textbf{Ablation on different method choices} Using a single tool call generally has worse performance compared to the multi-tool call variants. Score normalization has less of an effect.}
\label{tab:method_ablation}
\end{table*}


\begin{table}[htbp]
\centering
\small
\setlength{\tabcolsep}{5pt}
\begin{tabular}{lcccc}
\toprule
$\tau$ & LVB & VMME-all & M2M-Val & Mean \\
\midrule
2s  & 57.5 & 61.9 & \textbf{64.0} & 61.1 \\
5s  & 57.1 & 62.8 & 63.9 & 61.3 \\
10s & \textbf{61.8} & \textbf{64.6} & 62.0 & \textbf{62.8} \\
\bottomrule
\end{tabular}
\caption{\textbf{Effect of temporal gap $\tau$.} The optimal value of $\tau$ differs across datasets.}
\label{tab:tau}
\end{table}




\section{Conclusion}
We present ToolMerge, a keyframe retrieval method for long-video QA based on decomposition and merging: a planner decomposes the query into tool calls and specifies how their per-tool rankings are merged using boolean operators. To evaluate retrieval directly, we introduce Molmo-2 Moments (M2M), where each question is anchored to a specific time interval by construction.
Our experiments show that ToolMerge is competitive with prior keyframe selectors across different retrieval and question-answering tasks. We also show that the simple top-k SigLIP-2 baseline (SigLIP-Q) is a strong reference point across benchmarks. M2M's clip-anchored intervals also enable a recall-based reward, and GRPO on the M2M training split provides further improvement, especially on retrieval. Together, these results suggest that decomposition and merging is a useful framework for keyframe retrieval, especially for detailed caption retrieval. 
\paragraph{Acknowledgments} This research used both the DeltaAI advanced computing and data resource, which is supported by the National Science Foundation (award OAC 2320345) and the State of Illinois, and the Delta advanced computing and data resource which is supported by the National Science Foundation (award OAC 2005572) and the State of Illinois. Delta and DeltaAI are joint efforts of the University of Illinois Urbana-Champaign and its National Center for Supercomputing Applications. This research project has benefited from the Microsoft Agentic AI Research and Innovation (AARI) grant program. This work is supported in part by ONR award N00014-23-1-2383, NSF Grant 2519216, the DARPA Young Faculty Award, and ONR Grant N00014-26-1-2099.

Thanks to Savya Khosla for assisting with annotations. 
\bibliographystyle{plainnat}
\bibliography{bib}
\appendix
\section{Limitations}
\label{app:limitations}

A limitation of our work is that we focus on frame selection and do not address the answerer's reasoning abilities. We assume that given the right frames, a VLM produces the right answer, so questions whose difficulty lies in the reasoning step rather than in finding the relevant frames are not handled by our method. Downstream accuracy also reflects the VLM used (Qwen3-VL-8B, GPT-4o) and may differ with a different model. M2M questions are anchored to a single ground-truth interval and bounded by what the Molmo-2 caption describes, so queries distributed across the video or grounded in visual content the annotator did not narrate are out of scope. The tool set is fixed (SigLIP-2, T-REN, OCR), so the planner can only express decompositions these tools support, and capabilities outside this set, including audio understanding and fine-grained action recognition, are not covered.

\section{Broader Impacts}
\label{app:broaderimpact}
Returning keyframes alongside answers makes long-video QA systems
more inspectable: a user can audit which moments support a given
answer rather than trusting a text summary. The same retrieval
capabilities, however, can be applied to surveillance footage and
other privacy-sensitive video sources, where automated localization
of specific people, objects, or activities lowers the cost of
large-scale monitoring. We do not release new video data, and the
M2M benchmark is built on publicly available videos from the
Molmo-2 Captioning Dataset. Additionally, models trained on the data will inherit biases present in the videos and questions.
\section{Additional Results}
\label{sec:q25}
We compare ToolMerge with two additional keyframe selection methods that do not have publicly released code, MDP3~\cite{mdp3} and AIR~\cite{air}. AIR is a more expensive keyframe selection method since it requires iteratively calling a VLM to select frames. To matching the settings of the original papers, frames are sampled at 1 FPS and Qwen2.5-VL is the answering backbone. Otherwise, all settings match earlier experiments. As seen in Table~\ref{tab:qwen25}, ToolMerge is competitive with none-iterative based methods. 
\begin{table}[htbp]
\centering
\small
\setlength{\tabcolsep}{8pt}
\begin{tabular}{lcc}
\toprule
Method & Video-MME & LongVideoBench \\
\midrule
MDP3*       & 60.0 & 63.8 \\
ToolMerge  & 60.3 & 64.4 \\
AIR*  & 61.4 & 65.0 \\
\bottomrule
\end{tabular}
\caption{Comparison on Video-MME and LongVideoBench on Qwen2.5-VL with 32 frames at 1 FPS.*as reported in AIR~\citep{air}. ToolMerge is competitive on older backbones like Qwen2.5-VL especially with other keyframe selectoin works like MDP3. The more expensive AIR, which iteratively calls a VLM does slightly better.}
\label{tab:qwen25}
\end{table}

\section{Training Details}
\label{sec:training_details}
We discuss training details for the GRPO results in Section~\ref{sec:m2m}. The specific version of GRPO we use is DAPO~\citep{dapo} with Qwen3-VL-8B as the policy. We train with learning rate
$5{\times}10^{-6}$, 8 rollouts per prompt, batch size of 16 and 50 update steps. The vision encoder is frozen and so only the language model is updated. We use a 96 GB 4xH200  for training and the TRL library~\citep{trl}. 
\section{Compute Resources}
\label{sec:compute}
All experiments run on NVIDIA A100 40 GB GPUs for inference and H200 96 GB Gpus for training which typically takes 8 hours. Inference timing per component is reported in Table~\ref{tab:timing_breakdown}. Additional resources were used for preliminary experiments.


\section{Planner Prompt}
\label{app:prompt-planner}

The planner is a text-only LLM (Qwen3-VL-8B in the no-frame configuration) that receives the question, answer choices, video duration, and frame rate, and emits a set of tool queries combined by AND/OR. Tool queries target SigLIP-2 (visual similarity) and T-REN (object/entity detection); OCR runs unconditionally on every question and is not part of the planner's output.

\begin{lstlisting}
You are a search planner for a video question-answering system. Given a question and answer choices, write queries for specific search tools that LOCATE the relevant frames. A separate answerer model will look at those frames and determine the correct answer -- you do NOT answer the question yourself.

## Tools

**siglip** -- Visual similarity search.
  - Describe what the scene LOOKS LIKE: settings, actions, spatial layout, object attributes, visual states.
  - Cannot read text. Never include on-screen text in siglip queries.
  - Bad: "sign reading Exit Here" -> Good: "hallway with illuminated signs"
  - Bad: "someone is happy" -> Good: "person smiling and clapping"

**tren** -- Object and entity search.
  - Short noun phrases only. ONE entity per query.
  - Good at finding specific objects or people by appearance.
  - Bad: "person picking up a red mug from the table" -> Good: "red mug"
  - Bad: "cat and dog" -> Good: two separate queries, "cat" and "dog"

For both tools, focus on the most visually distinctive feature -- rare details beat generic descriptions.

## Query design

- Break complex scenes into separate queries across tools.
- Keep siglip queries concrete and visual. Avoid abstract or narrative language.
- Keep tren queries to short noun phrases. ONE entity per query.

## Combining queries (1-5 queries per plan)

- **AND** = intersection. Scene has multiple distinctive elements -- one query each. Never AND queries that describe the same thing differently.
- **OR** = union. Different scenes, or different queries that might each find what you need.

## OCR

OCR runs automatically on every question -- you do NOT need to handle it. Always write visual queries to locate the scene, even if the answer is about on-screen text or subtitles. The answerer model can read text directly from the frames you find.

## Rules

1. **Locate, don't answer.** Find the scene; the answerer decides what's happening.
2. **Always output at least one query.** Every question has a visual scene to find.
3. **Use all information.** Extract every visually searchable detail from the question AND the answer choices. Entities, objects, settings, actions -- if it can help locate the right frames, query for it.
4. **Use answer choices wisely.** Visually different choices -> search for each. Same scene described differently -> one query, let the answerer decide.
5. **Right tool:** siglip for scenes, actions, layout, visual states. tren for specific objects or people. Use both when you need both.

## Question

{question}

Options:
{options}

Video duration: {duration}s encoded at {fps} fps.

You MUST first write 1-3 sentences of reasoning before the JSON block. Think about: what must be visually true about the frames that contain the answer? What is the most distinctive element to search for? Do the answer choices point to different scenes or the same scene? Never output the JSON block without reasoning first.

Then output a JSON block:
```json
{"queries": [{"tool": "siglip", "query": "...", "id": "Q1"}], "combine": "Q1"}
```

Fields per query: "tool", "query", "id" (Q1, Q2, ...).

Examples:

---

Question: What does the woman in the red dress do after picking up the book from the table?
Options: A) places it on the shelf B) hands it to the man in glasses C) sits down on the couch and reads D) puts it in her bag E) walks out of the room

The question mentions a woman in a red dress, a book, and a table. The choices describe different actions after picking up the book -- each would look different visually. I'll find the woman, the book, and search for the distinct scenes from each choice.
```json
{"queries": [{"tool": "tren", "query": "woman in red dress", "id": "Q1"}, {"tool": "tren", "query": "book", "id": "Q2"}, {"tool": "siglip", "query": "person placing book on shelf", "id": "Q3"}, {"tool": "siglip", "query": "person handing book to someone", "id": "Q4"}, {"tool": "siglip", "query": "person sitting on couch reading", "id": "Q5"}], "combine": "(Q1 AND Q2) AND (Q3 OR Q4 OR Q5)"}
```

---

Question: What color is the vehicle that the man in the construction vest walks toward after crossing the street?
Options: A) red B) blue C) white D) black E) yellow

The question mentions a man in a construction vest and crossing a street. The choices are all vehicle colors -- visually distinct. I'll find the man and search for each colored vehicle near a street.
```json
{"queries": [{"tool": "tren", "query": "man in construction vest", "id": "Q1"}, {"tool": "siglip", "query": "person crossing street toward vehicle", "id": "Q2"}, {"tool": "tren", "query": "red vehicle", "id": "Q3"}, {"tool": "tren", "query": "blue vehicle", "id": "Q4"}, {"tool": "tren", "query": "white vehicle", "id": "Q5"}], "combine": "Q1 AND Q2 AND (Q3 OR Q4 OR Q5)"}
```

---

Question: In which room does the child first play with the wooden blocks?
Options: A) the kitchen B) the living room with the blue rug C) the bedroom D) the hallway E) the backyard

The question mentions a child and wooden blocks. The choices are different rooms, each visually distinct. I'll find the child and the blocks, and search for each room.
```json
{"queries": [{"tool": "tren", "query": "child", "id": "Q1"}, {"tool": "tren", "query": "wooden blocks", "id": "Q2"}, {"tool": "siglip", "query": "child playing in kitchen", "id": "Q3"}, {"tool": "siglip", "query": "living room with blue rug", "id": "Q4"}, {"tool": "siglip", "query": "child playing in bedroom", "id": "Q5"}], "combine": "(Q1 AND Q2) AND (Q3 OR Q4 OR Q5)"}
```

---

Question: What is shown on the display screen when the man in the blue jacket is standing at the podium?
Options: A) a bar chart B) a photo of the team C) the company logo D) a world map

The question asks about what appears on a display during a specific scene. OCR handles text automatically, so I need to find the scene visually -- the man in the blue jacket at the podium with a display screen.
```json
{"queries": [{"tool": "tren", "query": "man in blue jacket", "id": "Q1"}, {"tool": "siglip", "query": "person standing at podium with display screen", "id": "Q2"}], "combine": "Q1 AND Q2"}
```

---

Question: What is the name of the restaurant shown on the sign outside the building?
Options: A) Mario's B) The Golden Fork C) Sushi Palace D) Burger Barn E) Cafe Luna

The question asks about text on a sign outside a building. I need to find the building exterior with the sign -- the answerer will read the text.
```json
{"queries": [{"tool": "siglip", "query": "building exterior with restaurant sign", "id": "Q1"}], "combine": "Q1"}
```
\end{lstlisting}

\section{Benchmark Construction Prompts}
\label{app:prompts}

We document the substantive prompts used in the benchmark construction pipeline (Section~\ref{sec:dataset_ant}). Boilerplate ``answer with the letter A--E'' prompts used by the blind-LLM filter (Step~2), necessity filter (Step~6), and difficulty filter (Step~7) are omitted. Placeholders of the form \texttt{\{name\}} are substituted at runtime.

\subsection{Step 1: Question Generation}
\label{app:prompt-genq}

Model: GPT-5.4. Given a clip caption and an optional video summary, generate 5-option multiple-choice questions.

\paragraph{System prompt.}
\begin{lstlisting}
You are an expert at creating challenging multiple-choice questions for a video understanding benchmark.

You will receive two pieces of context:
1. A VIDEO SUMMARY describing the full video at a high level.
2. A CLIP CAPTION describing a specific segment of that video in detail.

Your job is to generate as many as possible high-quality questions about the clip that test whether someone actually watched the video. The viewer has access to the ENTIRE video, not just the clip -- so frame questions naturally, as if asking someone who watched the whole thing.

Use the video summary only for context about the setting, topic, and overall structure. All question answers must come from specific visual details in the clip caption -- never generate a question whose answer requires only the video summary.

Question types to generate (generate MULTIPLE questions per type when the caption supports it):

SCENE CO-OCCURRENCE -- Ground the question in a specific visible moment ("When X is happening / is visible, what else is on screen?"). The anchor must be a concrete, specific visual detail -- not a vague reference like "in the scene" or "at one point." The answer choices should be detailed descriptions, not single words. Generate one for each distinct moment the clip caption describes in detail.

SPATIAL RELATIONS -- Ask about the position or arrangement of elements relative to each other. The question must specify WHICH scene or moment you are asking about using a concrete visual anchor ("In the shot where the man holds up the red jar, where is the cutting board relative to the stove?"). Never ask unanchored spatial questions like "Where is the man?" or "What is on the left side?"

CROSS-REFERENCING -- Combine multiple details from different moments to ask a question that requires tracking an element across the clip. For example: an object that appears in two different configurations, a person who moves between locations, or an item that is used for different purposes at different points.

VISUAL DETAIL -- Ask about a specific, precise visual detail that is easy to miss: a color, a label, a count, a texture, a gesture, a facial expression, a piece of clothing, or a small object. The question must be anchored to a specific moment ("While X is doing Y, what color is Z?"). These questions reward careful observation.

GROUNDING RULES (critical):
- Every question MUST specify the moment or visual context it refers to. Use concrete anchors: a specific action being performed, a specific object visible on screen, a specific person doing a specific thing.
- BAD (too vague): "Where is the woman standing?" / "What is on the table?" / "What does the man do?"
- GOOD (grounded): "When the woman in the green apron lifts the lid off the pot, where is she standing relative to the kitchen island?" / "In the shot where three bottles are visible on the table next to the notebook, what color is the middle bottle?" / "After the man sets down the wrench and picks up the tape measure, what does he do with his other hand?"
- If a question could apply to multiple moments in the video, it is too vague. Add specificity until it pins down exactly one moment.

DISTRACTOR RULES (critical -- follow these carefully):
- Every wrong answer must be plausible within the same scene type and setting. If the clip takes place in a gym, all wrong answers must involve gym-relevant activities, equipment, or body positions. If the clip is a cooking scene, wrong answers must involve cooking-relevant actions and tools. NEVER use scenarios from unrelated domains.
- The best wrong answers take a true detail and change one specific aspect: swap a color, swap which hand or body part is used, swap the relative positions of two objects, swap a count, or attribute a detail to the wrong moment. A wrong answer should feel like a misremembering of what actually happened.
- For co-occurrence questions, pair real elements from the clip with the wrong moment, or describe a plausible element that fits the setting but isn't actually present.
- A viewer who watched the video carelessly or only partially should find at least 2-3 wrong answers tempting. If a wrong answer is obviously ridiculous to someone who has never seen the video, it is a bad distractor -- rewrite it.

ADVERSARIAL SELF-CHECK (required -- apply to every question before outputting):
After drafting each question and its 5 choices, check: could someone who has NEVER seen this video -- relying only on common sense and world knowledge -- identify the correct answer? Common giveaways to fix:
- The correct answer is longer or more detailed than the wrong answers
- The correct answer is the only one that "makes sense" given the setting (e.g., everyone knows kitchens have stoves -- so "near the stove" is guessable without watching)
- The correct answer uses specific language from the question while wrong answers are vague
- World knowledge alone makes one answer obvious (e.g., "the chef adds salt" in a cooking video)
- The wrong answers contradict themselves or describe impossible actions
- The wrong answers all share a pattern that the correct answer breaks (e.g., four wrong answers mention the left side and the correct answer says right)
If any of these apply, rewrite until the correct answer is NOT distinguishable without watching the video. All 5 choices must match in length, specificity, and plausibility.

FORBIDDEN CONTENT -- Do NOT generate questions about:
- Camera movement, angles, zoom, or shot composition
- Audio, narration, music, sound effects, or dialogue content
- The "description," "caption," "summary," or "text" -- never reference these words
- Temporal ordering (which event came first/last, correct sequence of events)
- Meta-questions like "What is this video about?" or "What is the purpose of the video?"

ANSWER FORMAT:
- Generate exactly 1 correct answer and exactly 4 wrong answers per question, labeled (A) through (E)
- The correct answer can be in ANY position -- distribute it roughly evenly across A-E over all questions
- All 4 wrong answers must be independently plausible and follow the distractor rules above
- Answer choices should be descriptive phrases or sentences, not single words
- Every question must be answerable ONLY from the clip caption content -- do not invent details beyond what is described
- Exactly ONE correct answer -- no two choices should be synonyms or both defensible
- Questions must read as if asked to a video viewer. Never reference "the caption", "described", "mentioned", or "the text" -- phrase everything as "in the video", "visible in the clip", "shown on screen", etc.
- Generate 3-5 high-quality questions. Prioritize diversity across question types. Prefer fewer excellent questions over many mediocre ones.

Format each question like this:

Q1: [question text]
(A) [choice A]
(B) [choice B]
(C) [choice C]
(D) [choice D]
(E) [choice E]
Answer: [letter]

Q2: [question text]
...
\end{lstlisting}

\paragraph{User template.}
\begin{lstlisting}
Video summary:
{video_summary}

Clip caption:
{clip_caption}

Generate 3-5 multiple-choice questions from this clip caption.
\end{lstlisting}

\subsection{Step 3: Answer-Choice Rewrite}
\label{app:prompt-rewrite}

Model: GPT-5.2. Triggered when the blind-LLM filter (Step~2) scores at or above the discard threshold. Three sequential calls per question; the rewritten options are then re-checked by Step~2.

\paragraph{Call 1 -- generate plausible candidates.}
System:
\begin{lstlisting}
You are generating wrong answers for a video understanding benchmark. Focus ONLY on making plausible, hard-to-eliminate distractors. Do NOT worry about matching the length or format of the correct answer -- that will be handled separately.
\end{lstlisting}

User:
\begin{lstlisting}
Question: {question}
Correct answer: {correct_text}

Generate exactly 6 plausible wrong answers that:
1. Are plausible for the question topic -- swap specific details (color, direction, object, name, count, position)
2. Would be tempting to someone who watched the video carelessly or only partially
3. Are clearly wrong (not synonyms or paraphrases of the correct answer)
4. Do NOT all share a pattern that the correct answer breaks
5. Cover diverse alternatives -- don't just vary one detail across all 6

Return ONLY valid JSON (no markdown fences):
{"wrong_answers": ["...", "...", "...", "...", "...", "..."]}
\end{lstlisting}

\paragraph{Call 2 -- analyze correct-answer format.}
System:
\begin{lstlisting}
You are a text format analyst. Analyze the format of the given answer choice precisely.
\end{lstlisting}

User:
\begin{lstlisting}
Given this correct answer for a multiple-choice question, describe its format precisely:
Answer: "{correct_text}"

Return ONLY valid JSON (no markdown fences):
{"syntactic_form": "noun phrase / verb phrase / full sentence / etc",
 "word_count": N,
 "structure": "e.g. article + adjective + noun + prepositional phrase",
 "specificity": "single-detail / multi-detail / enumeration"}
\end{lstlisting}

\paragraph{Call 3 -- reformat candidates to match.}
System:
\begin{lstlisting}
You are reformatting answer choices for a multiple-choice benchmark. Your job is to rewrite each wrong answer so it matches the correct answer's format EXACTLY -- same syntactic form, similar word count, same structure pattern -- while preserving the original meaning.
\end{lstlisting}

User:
\begin{lstlisting}
Correct answer: {correct_text}

The correct answer has this format:
- Syntactic form: {syntactic_form}
- Word count: {word_count}
- Structure: {structure}
- Specificity: {specificity}

Rewrite each of these wrong answers to match that format exactly (+/- 2 words), preserving their meaning:
1. {wrong_1}
2. {wrong_2}
3. {wrong_3}
4. {wrong_4}

Return ONLY valid JSON (no markdown fences):
{"reformatted": ["...", "...", "...", "..."]}
\end{lstlisting}

\subsection{Step 4: Scope Filter}
\label{app:prompt-scope}

Model: GPT-5.2 (temperature 0). Two independent passes; failing either discards the question.

\paragraph{4a. Forbidden content.}
\begin{lstlisting}
You are a strict quality auditor for a video-understanding benchmark. You will receive a MULTIPLE-CHOICE QUESTION with 5 answer choices (A-E) and the correct answer letter.

Your ONLY task: check if the question contains FORBIDDEN CONTENT.

Flag the question as FAIL if it asks about ANY of the following:
- Camera work (movement, angles, zoom, panning, composition, framing)
- Audio, narration, music, dialogue, sound effects, voiceover
- References to "description", "caption", "summary", or "text"
- Meta-questions ("What is this video about?", "What is the main topic?")
- Single-frame language like "in the shot" or "in this shot" (questions should reference video moments, not individual shots)

Return ONLY valid JSON (no markdown fences):
{"pass": true}
or
{"pass": false}
\end{lstlisting}

\paragraph{4b. Out-of-clip evidence.}
\begin{lstlisting}
You are a strict quality auditor for a video-understanding benchmark. You will receive:
- Optionally, a VIDEO SUMMARY describing the full video at a high level.
- A CLIP CAPTION describing a specific segment of that video in detail.
- A MULTIPLE-CHOICE QUESTION with 5 answer choices (A-E) and the correct answer letter.

Your ONLY task: check if the question requires knowledge OUTSIDE this clip.

Flag the question as FAIL if:
- It references "beginning/end of video" when the clip is from the middle
- It asks what comes before/after events at clip boundaries
- It spans beyond what the caption describes
- Any answer choice references events not described in the caption

Within-clip temporal relations are fine (e.g., "What happened before X?" when both events are in the caption).

Return ONLY valid JSON (no markdown fences):
{"pass": true}
or
{"pass": false}
\end{lstlisting}

\subsection{Step 5: Cleanup Rewrite}
\label{app:prompt-cleanup}

Four sequential passes. Each returns \texttt{\{"changed": false\}} or supplies rewritten fields, which are applied in place.

\paragraph{5a. Answer-choice cleanup (GPT-5.2).}
\begin{lstlisting}
You are a strict quality auditor for a video-understanding benchmark. You will receive:
- Optionally, a VIDEO SUMMARY describing the full video at a high level.
- A CLIP CAPTION describing a specific segment of that video in detail.
- A MULTIPLE-CHOICE QUESTION with 5 answer choices (A-E) and the correct answer letter.

Your ONLY task: check the ANSWER CHOICES for problems and rewrite if needed.

Problems to fix:
- Two choices overlap or are both correct
- Wrong answers are implausible or from the wrong domain
- Correct answer stands out by length, detail, or pattern
- Correct answer is the only "sensible" choice given world knowledge (text-answerable)

If you find problems, rewrite the choices to fix them while keeping the correct answer faithful to the caption. Preserve the correct answer letter.

Return ONLY valid JSON (no markdown fences):
{"changed": false}
or
{"changed": true, "rewritten_options": {"A": "...", "B": "...", "C": "...", "D": "...", "E": "..."}}
\end{lstlisting}

\paragraph{5b. Grounding (GPT-5.2).}
\begin{lstlisting}
You are a strict quality auditor for a video-understanding benchmark. You will receive:
- Optionally, a VIDEO SUMMARY describing the full video at a high level.
- A CLIP CAPTION describing a specific segment of that video in detail.
- A MULTIPLE-CHOICE QUESTION with 5 answer choices (A-E) and the correct answer letter.

Your ONLY task: check if the question is properly GROUNDED to a specific moment.

The question must anchor to ONE specific moment or visual detail (e.g., "When the person picks up the red cup" or "After the car turns left at the intersection").

Fails if the question is vague ("in the scene", "in the video") or ambiguous when the caption describes repeated similar events.

If the question lacks grounding, rewrite it to add a concrete visual anchor from the caption. The rewritten question can be long -- include as much descriptive detail as needed to uniquely identify the moment (e.g., "When the woman in the red jacket crouches down near the wooden fence and reaches toward the small dog, what does she pick up?"). Do not sacrifice specificity for brevity.

Return ONLY valid JSON (no markdown fences):
{"changed": false}
or
{"changed": true, "rewritten_question": "..."}
\end{lstlisting}

\paragraph{5c. Proper names $\rightarrow$ visual descriptions (GPT-5.2).}
\begin{lstlisting}
You are a strict quality auditor for a video-understanding benchmark. You will receive:
- Optionally, a VIDEO SUMMARY describing the full video at a high level.
- A CLIP CAPTION describing a specific segment of that video in detail.
- A MULTIPLE-CHOICE QUESTION with 5 answer choices (A-E) and the correct answer letter.

Your ONLY task: replace any PERSON NAMES with visual descriptions.

Replace any person's proper name (e.g., "John", "Dr. Smith", "Sarah") with a specific visual description from the caption (e.g., "the man in the blue shirt", "the woman with the red hat").

If the caption provides specific visual details about the person, USE THEM to create a distinctive description. Do not use generic descriptions like "a person" or "the man" when more specific details are available in the caption.

Apply to both the question text and all answer choices. Place names, brand names, and object names are fine -- only replace person names.

Return ONLY valid JSON (no markdown fences):
{"changed": false}
or
{"changed": true, "rewritten_question": "...", "rewritten_options": {"A": "...", "B": "...", "C": "...", "D": "...", "E": "..."}}

Only include "rewritten_question" if the question text changed. Only include "rewritten_options" if any option text changed. Include both if both changed.
\end{lstlisting}

\paragraph{5d. Caption-revealing language (GPT-4.1-mini, temperature 0).}
\begin{lstlisting}
You are a strict quality auditor for a video-understanding benchmark. You will receive a MULTIPLE-CHOICE QUESTION with 5 answer choices (A-E) and the correct answer letter.

Your ONLY task: remove CAPTION-REVEALING LANGUAGE.

Rewrite any phrases that reveal the question was generated from a caption or text description:
- "is described as" -> "is" or "appears to be"
- "is shown to be" -> "is"
- "could be described as" -> "could be" or "appears"
- "according to the description" -> remove entirely
- Similar meta-phrasing that references a written source

The question should read as if written by someone watching the video, not reading a description.

Apply to both the question text and all answer choices.

Return ONLY valid JSON (no markdown fences):
{"changed": false}
or
{"changed": true, "rewritten_question": "...", "rewritten_options": {"A": "...", "B": "...", "C": "...", "D": "...", "E": "..."}}

Only include "rewritten_question" if the question text changed. Only include "rewritten_options" if any option text changed. Include both if both changed.
\end{lstlisting}

\subsection{Step 8: Diversity Sampling}
\label{app:prompt-dedup}

Model: GPT-4.1-nano (temperature 0).

\begin{lstlisting}
The following {n} questions all reference the same short video clip.

Pick UP TO {cap} questions that together give the most diverse, non-redundant
coverage of the clip. Two questions are redundant if they probe the same
specific detail even when worded differently -- keep only one of them.

RETURN FEWER than {cap} if any additional choice would duplicate one already
chosen. There is no penalty for returning fewer.

Questions:
{questions_text}

Output format: a single line containing the chosen question numbers separated
by commas, nothing else. Example: `1, 4, 7`
Use only numbers that appear in brackets above (1 through {n}).
\end{lstlisting}




\end{document}